\documentclass[letterpaper, 10 pt, conference]{ieeeconf}  

\IEEEoverridecommandlockouts                              

\usepackage{cite}
\usepackage[pdftex]{graphicx}
\usepackage[caption=false,font=footnotesize]{subfig}

\usepackage{eso-pic}
\usepackage{url}
\AddToShipoutPicture*{%
     \AtTextUpperLeft{%
         \put(-3.5,10){
           \begin{minipage}{\textwidth}
              \scriptsize
              \MakeUppercase{Preprint version of an IEEE Robotics and Automation Letters article; final version available at} \url{https://ieeexplore.ieee.org/Xplore}
           \end{minipage}}%
     }%
}

\usepackage{xspace}
\makeatletter
\DeclareRobustCommand\onedot{\futurelet\@let@token\@onedot}
\def\@onedot{\ifx\@let@token.\else.\null\fi\xspace}

\def\eg{\emph{e.g}\onedot} 
\def\ie{\emph{i.e}\onedot}

\def\etal{\emph{et al}\onedot}

\usepackage{url}
\usepackage{graphicx}

\usepackage{tabularx}
\usepackage{multirow}
\usepackage[nopatch=eqnum]{microtype}
\usepackage{booktabs}      
\usepackage[table]{xcolor} 
\let\labelindent\relax
\usepackage{enumitem}
\usepackage{ifthen}

\begin{document}
\bstctlcite{bstctl:forced_etal,bstctl:nodash}

\newif\ifAUTHORS
\AUTHORStrue 

\title{\LARGE \bf
Automated Coral Spawn Monitoring for Reef Restoration:\\The Coral Spawn and Larvae Imaging Camera System (CSLICS)}

\ifAUTHORS
\author{Dorian Tsai$^{1}$, 
    Christopher A. Brunner$^{2}$, 
    Riki Lamont$^{1,2}$, 
    F. Mikaela Nordborg$^{2}$, 
    Andrea Severati$^{2}$, \\
    Java Terry$^{1}$, 
    Karen Jackel$^{1}$,
    Matthew Dunbabin$^{1}$, 
    Tobias Fischer$^{1}$ and 
    Scarlett Raine$^{1}$
\thanks{$^{1}$D.T., R.L., J.T., K.J., M.D., T.F.~and S.R.~are with the Queensland University of Technology (QUT), Brisbane, Australia and acknowledge continued support through the Centre for Robotics. {\tt\small dy.tsai@qut.edu.au}}
\thanks{$^{2}$C.A.B., R.L.~(CSLICS work conducted at QUT), F.M.N. and A.S. are with the Australian Institute for Marine Science (AIMS), Townsville, Australia.}%
\thanks{The authors acknowledge the Manbarra, Wulgurukaba and Bindal people as the Traditional Owners of the land and sea Country where this research was performed. Adult corals and spawn were collected from the sea Country of the Manbarra and Bindal peoples and used with their Free Prior and Informed Consent. The authors are grateful for the granting of consent, acknowledge Traditional Owners as the first scientists and custodians, and pay respects to Elders, past, present and emerging.
We acknowledge QUT Research Engineering Facilities for design and engineering support, QUT Design and Fabrication Services for manufacturing work, the AIMS Technology Transformation team for engineering support, technology scaling and operationalization, and the AIMS National Sea Simulator (SeaSim) team for their facilities and coral husbandry work the Reef Restoration and Adaptation Program (RRAP), which is funded by a partnership between the Australian Government’s Reef Trust and the Great Barrier Reef Foundation. T.F.~acknowledges funding from an Australian Research Council Discovery Early Career Researcher Award Fellowship DE240100149.}%
}
\else
\author{Anonymous Submission}
\fi

\maketitle
\thispagestyle{empty}
\pagestyle{empty}

\begin{abstract}

Coral aquaculture for reef restoration requires accurate and continuous spawn counting for resource distribution and larval health monitoring, but current methods are labor-intensive and represent a critical bottleneck in the coral production pipeline. 
We propose the Coral Spawn and Larvae Imaging Camera System (CSLICS), which uses low cost modular cameras and object detectors trained using human-in-the-loop labeling approaches for automated spawn counting in larval rearing tanks. This paper details the system engineering, dataset collection, and computer vision techniques to detect, classify and count coral spawn. 
Experimental results from mass spawning events demonstrate an F1 score of 82.4\% for surface spawn detection at different embryogenesis stages, 83\% F1 score for sub-surface spawn detection, and a saving of 5,720 hours of labor per spawning event compared to manual sampling methods at the same frequency. Comparison of manual counts with CSLICS monitoring during a mass coral spawning event on the Great Barrier Reef demonstrates CSLICS' accurate measurement of fertilization success and sub-surface spawn counts. 
These findings enhance the coral aquaculture process and enable upscaling of reef restoration efforts to address climate change threats facing ecosystems like the Great Barrier Reef.
\end{abstract}

\section{Introduction}
\label{sec:intro}

The Great Barrier Reef (GBR) is one of the world's most important and diverse ecosystems, serving as a habitat for over 25\% of all marine species and as a nursery for over 75\%~\cite{heron2017impacts}. Coral reefs also provide social, economic, and cultural services with an estimated value of over 1 trillion USD globally~\cite{heron2017impacts, costanza2014changes}. However, coral reefs around the world are severely threatened by climate change~\cite{hughes2017global}. Prolonged exposure to abnormally high ocean temperatures causes mass coral bleaching and death, resulting in a projected 70-90\% decrease in live coral on reefs by 2050 if no action is taken~\cite{de201227, Souter2021Status}. If coral reefs are to survive these accelerating pressures, scalable and effective restoration approaches are urgently required.


Coral aquaculture has emerged as one such approach, enabling large-scale sexual reproduction of corals in controlled facilities~\cite{randall2020sexual}. Unlike asexual fragmentation-based methods, sexual reproduction can produce orders of magnitude more corals while promoting genetic diversity and adaptive capacity to environmental change~\cite{randall2020sexual, banaszak2023applying, dela2020enhancing}. The long-term success of coral aquaculture, however, depends critically on monitoring and managing the millions of early-stage coral spawn and larvae in culture tanks. Automated and continuous monitoring is therefore essential to ensure that this otherwise scalable method delivers viable recruits at the scale required for meaningful reef restoration.

\begin{figure}[t]
    \centering
    \subfloat{\includegraphics[width=0.532\columnwidth, clip, trim={9.5cm 2cm 10.5cm 2cm}]{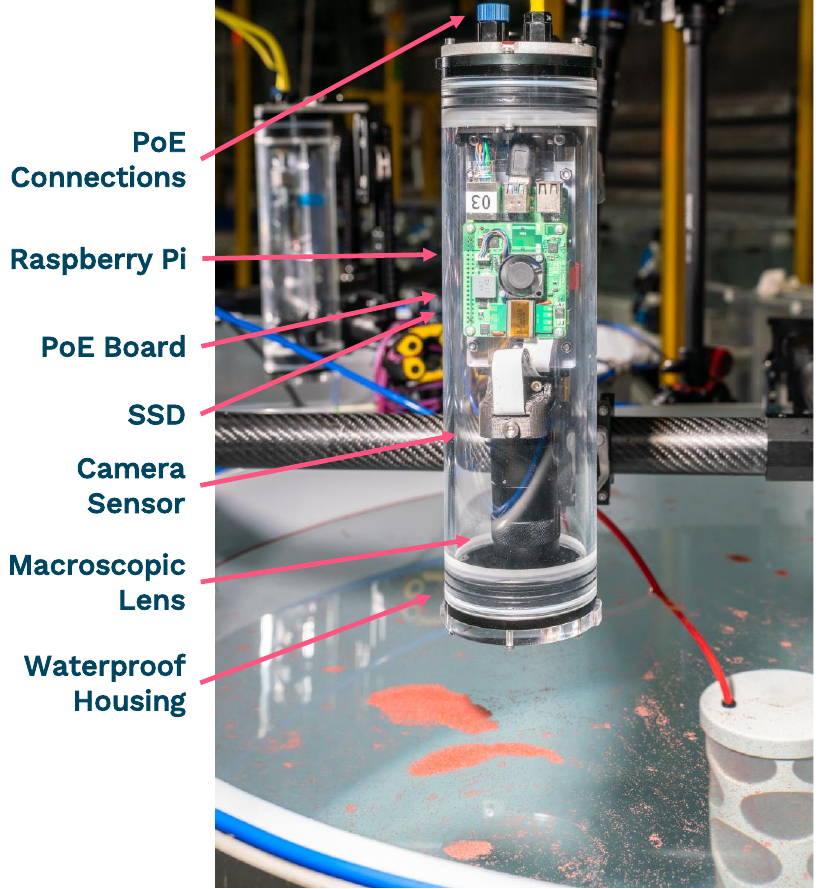}} \hfill
    \subfloat{\includegraphics[width=0.45\columnwidth]{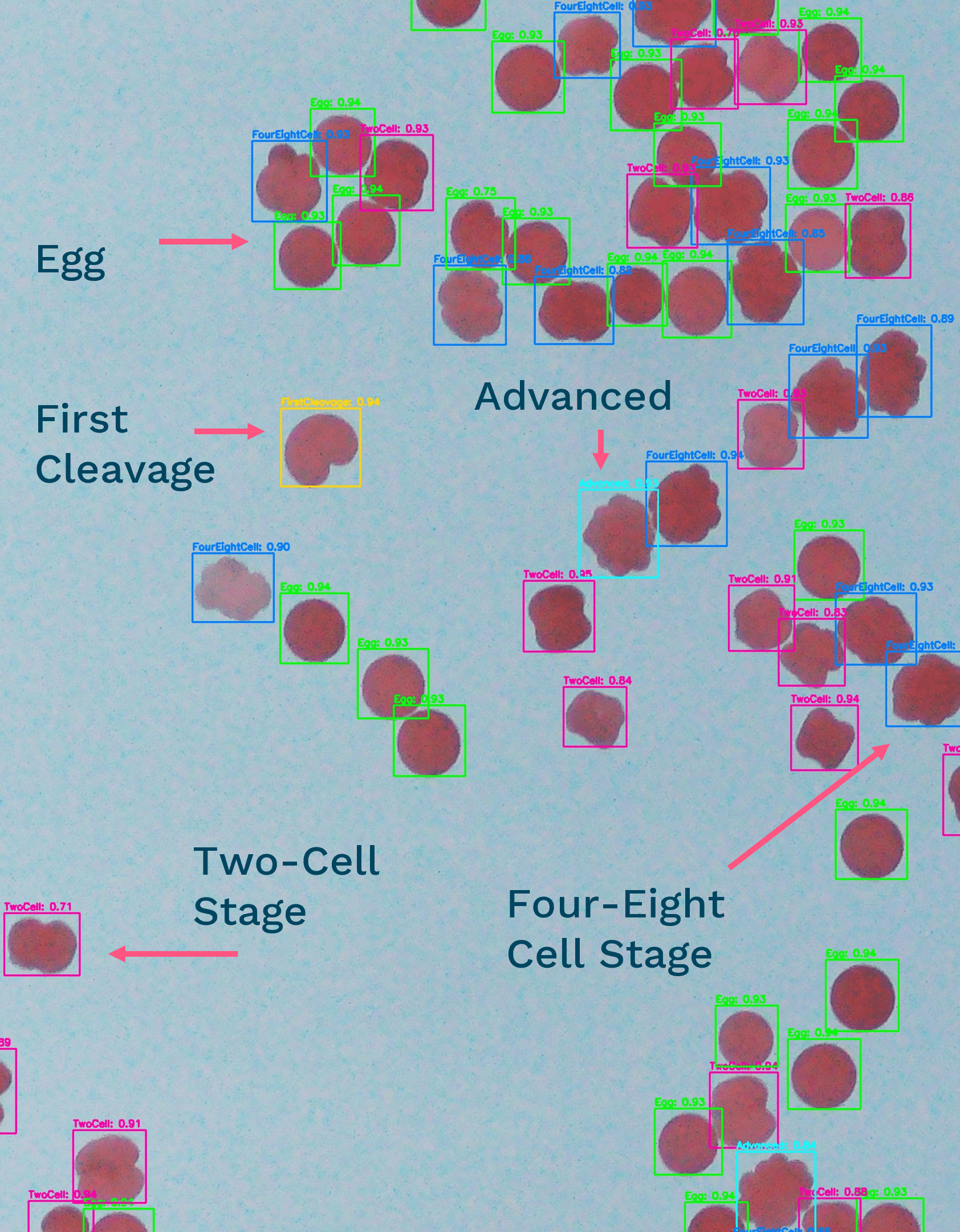}}
    \caption{The Coral Spawn and Larval Imaging Camera System (CSLICS) mounted above the larval rearing tank during surface monitoring with the coral spawn floating inside (left). Sample CSLICS detections highlight the developmental progression of coral embryogenesis, from unfertilized eggs, followed by first cleavage (the first confirmation of fertilization), then two-cell, four-to-eight and advanced cell stages (right).}
    \vspace{-0.3cm}
    \label{fig:cslics}
\end{figure}

The coral aquaculture process involves broodstock selection, spawn collection, fertilization and larval rearing, settlement, and reseeding to reefs~\cite{pollock2017coral, raine2025ai}. A critical bottleneck lies in larval rearing, where millions of fragile eggs and embryos must be monitored during their early life stages~\cite{banaszak2023applying}. Quantitative measures, such as fertilization success, are vital for predicting culture viability, yet current monitoring relies on manual sampling and visual counts under a microscope—a process which takes up to 20 minutes per sample. With large facilities projected to operate over 60 rearing tanks by 2030, manual monitoring is infeasible: operators are often limited to sampling only a subset of tanks once per day. This data sparsity, combined with the fact that cultures can deteriorate within hours without intervention, makes manual monitoring both labor-intensive and inadequate for large-scale restoration.


To overcome this monitoring bottleneck, we propose the Coral Spawn and Larvae Imaging Camera System (CSLICS), shown in Fig.~\ref{fig:cslics}. CSLICS is a submersible, tank-mounted microscopic camera system designed to automatically capture and analyze images of developing coral during their early life stages. By leveraging computer vision techniques for spawn detection and counting, CSLICS reduces manual handling risks while enabling continuous, non-invasive monitoring at high temporal resolution. A human-in-the-loop, semi-supervised annotation strategy further streamlines dataset development across different embryogenesis\footnote{The process by which a fertilized egg develops into an embryo, which is the first stage of development for a multicellular organism.} stages, ensuring robust detector performance with limited labeling effort. CSLICS enables detailed and responsive monitoring, and can provide per-image counts every 10 seconds.  When deployed during a mass coral spawning event on the GBR, CSLICS was used to sample every five minutes during the surface phase (fertilization), and every hour during the sub-surface phase (late embryogenesis and larval stages). Continuous analysis increases the temporal resolution of culture health data beyond what is feasible with manual methods, and may also enable identifying correlations between culture health/quality and factors such as tank or environmental conditions, and manual handling processes.  

Our contributions are as follows:
\begin{enumerate}
    \item For the first time, we propose the design of a tank-mounted computer vision-based coral spawn monitoring system for coral aquaculture (Fig.~\ref{fig:cslics}).
    \item We develop and practically integrate a coral spawn detector and counting algorithm for two commonly occurring, reef-building coral species in the Indo-Pacific, \textit{Acropora kenti} and \textit{Acropora loripes}, which enables automatically measuring and assessing coral fertilization success.
    \item We demonstrate the viability of automated, low-cost coral monitoring technologies, by reporting results during their deployment in a state-of-the-art, large-scale coral aquaculture research facility, where CSLICS provides scientists with real-time monitoring data to support mass coral production during annual spawning events.
\end{enumerate}
We will make our trained models and code publicly available upon acceptance to foster future research.


\begin{figure}[t]
    \centering
    \ifAUTHORS
    \subfloat{\includegraphics[height=0.485\columnwidth, width=0.35\columnwidth]{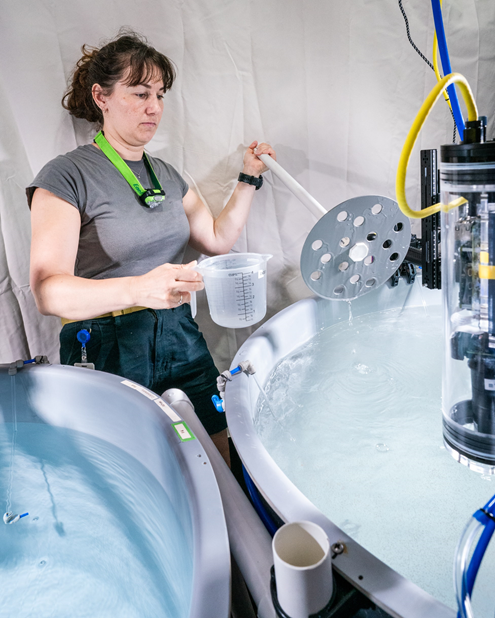}} \hfill
    \subfloat{\includegraphics[height=0.485\columnwidth, width=0.64\columnwidth]{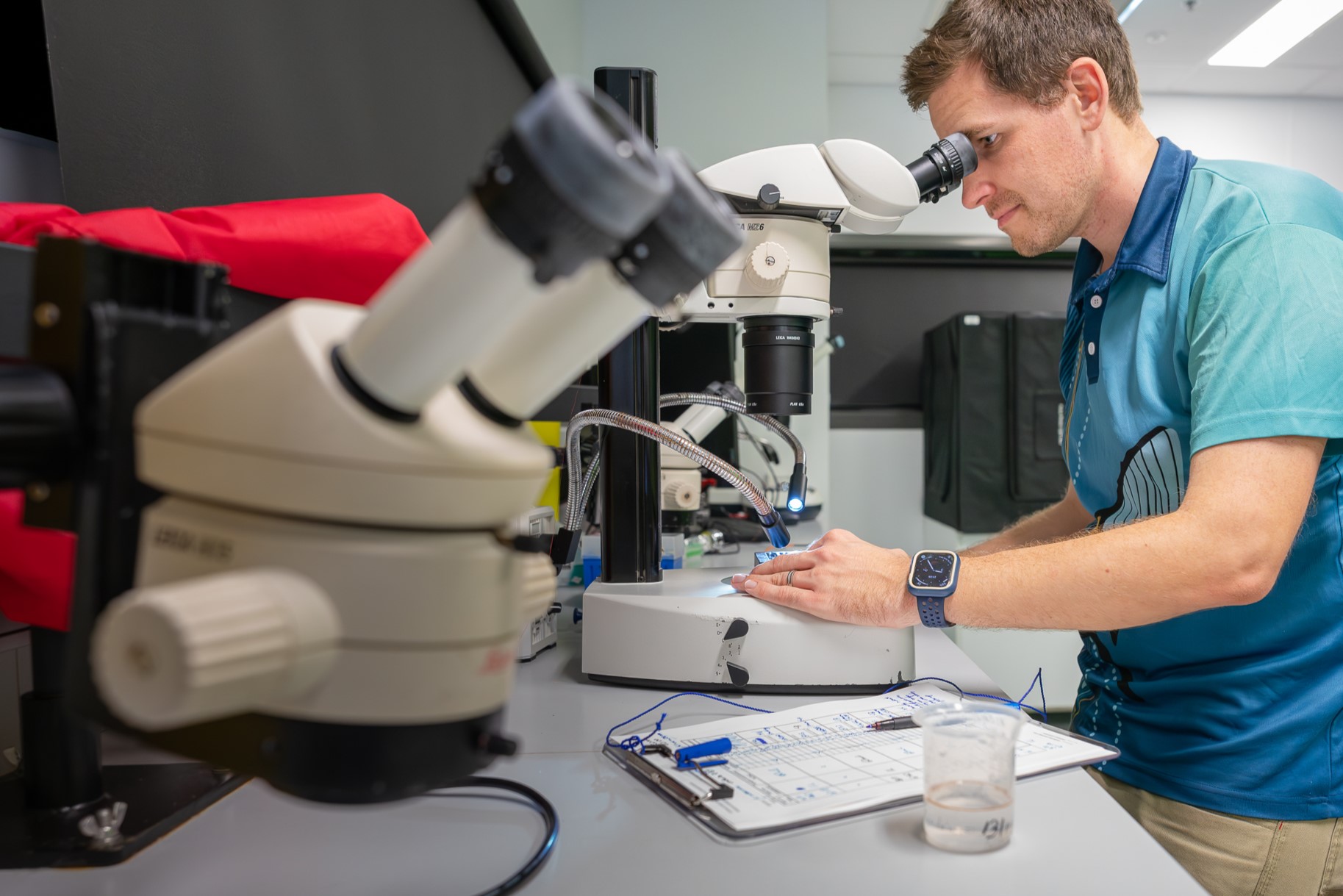}}
    \else
    \subfloat{\includegraphics[height=0.485\columnwidth, width=0.35\columnwidth]{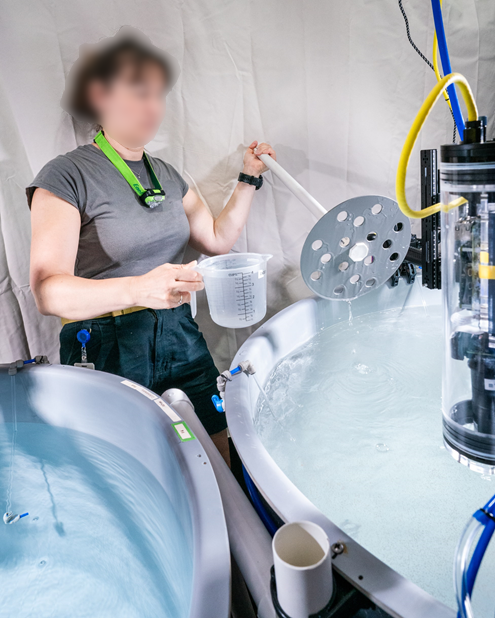}} \hfill
    \subfloat{\includegraphics[height=0.485\columnwidth, width=0.64\columnwidth]{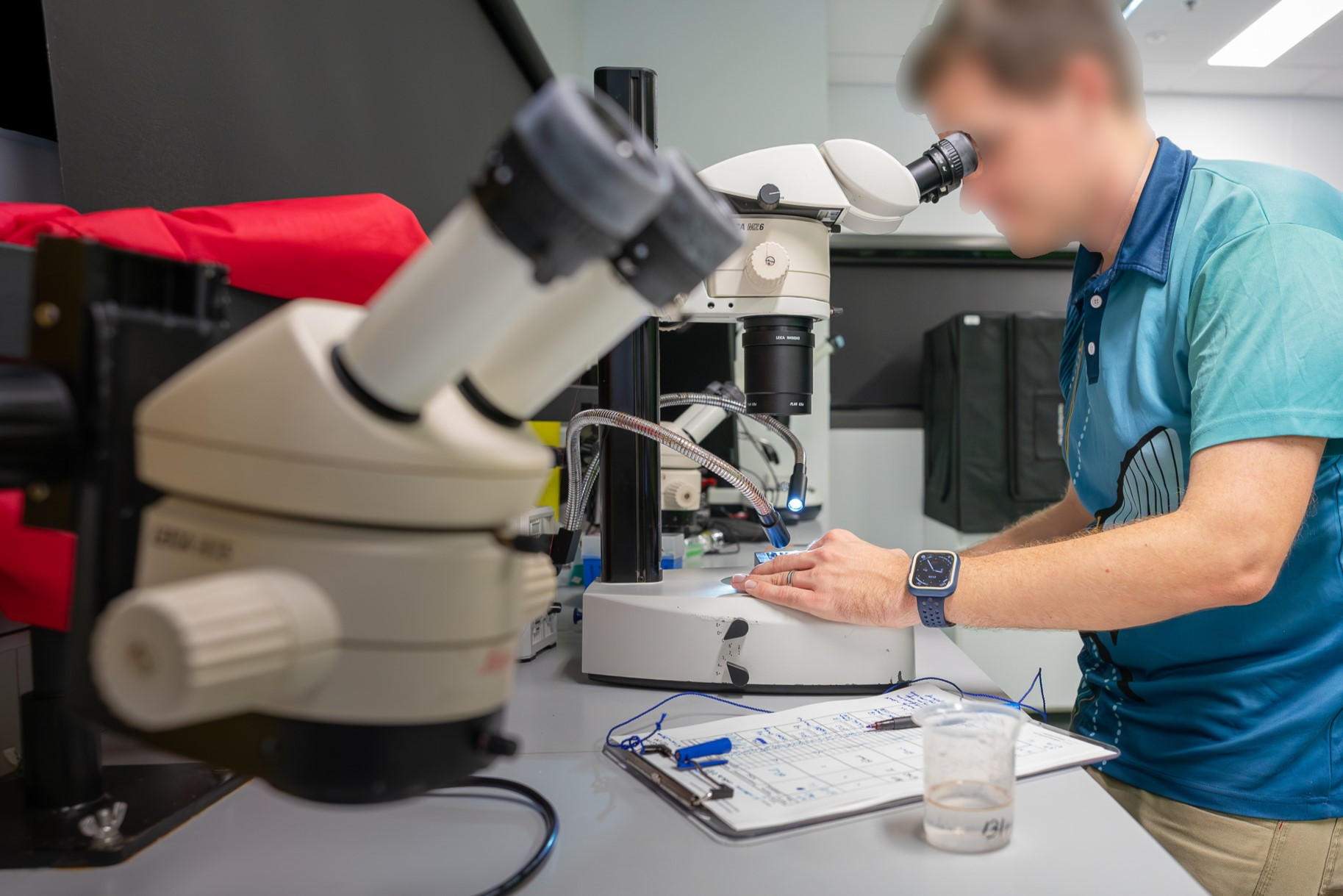}}
    \fi   
    \caption{Current methods of counting coral spawn are intensely time-consuming and manual, involving stirring the larval culture to homogenize the larvae and sampling from the tanks multiple times (left), and then counting individual corals with a stereo microscope (right).}
    \label{fig:manual_counting}
    \vspace{-0.4cm}
\end{figure}


\section{Related Work}
\label{sec:related}

Monitoring coral spawn traditionally relies on manual observation and counting, a process that is time-consuming, vulnerable to human error and disruptive to the delicate coral spawn. There have been some recent works exploring the use of technology at different stages of the coral production process~\cite{severati2024autospawner, raine2025ai}, and recent advances in deep learning have further expanded the potential for automated monitoring, however existing approaches are not designed specifically for coral spawn monitoring at the early, free-swimming \mbox{stage -- a} key bottleneck in coral production. This section first describes the process of manual coral spawn monitoring, reviews advances in automated coral spawn monitoring, and outlines the application of deep learning for counting tasks and annotating imagery.

\begin{figure*}[t]
\centering
\includegraphics[width=\linewidth,clip,trim=4.2cm 3.8cm 4.2cm 3.8cm]{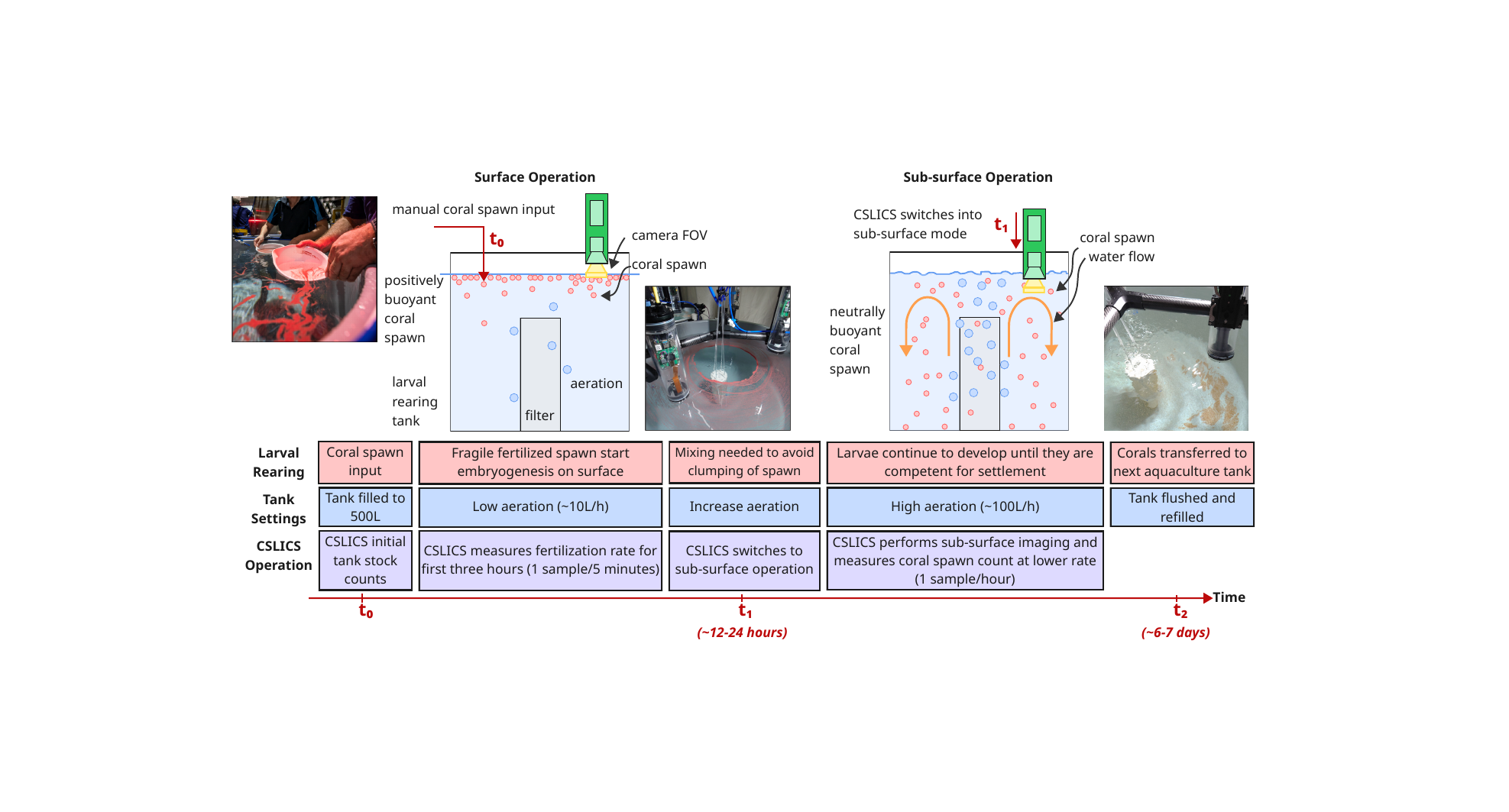}
\vspace*{-0.3cm}
\caption{CSLICS has two operational modes for different stages of spawn development: a) for the first 12-24 hours (t\textsubscript{0}--t\textsubscript{1}), CSLICS is in surface operation, and after t\textsubscript{1} it switches to b) sub-surface operation (t\textsubscript{1}--t\textsubscript{2}). The larval rearing process is described in red, with the parallel tank settings made by an operator (shown in blue), and the simultaneous CSLICS operations (in purple).}
\label{fig:cslicsmodes}
\vspace{-0.4cm}
\end{figure*}

\subsection{Manual Coral Spawn Monitoring}

In marine science, coral spawn counting is predominantly performed manually due to the complexity of the task, the delicate nature of the developing coral larvae, and the typically small scale of operations~\cite{randall2020rapid}. During the critical annual mass-spawning events, consistency in counting is essential, yet manual methods can vary based on the sampling process and the individual performing the counts.

Manual spawn counting in coral aquaculture involves sampling from the tank and then counting from the samples, as illustrated in Fig.~\ref{fig:manual_counting}. Therefore, traditional sampling requires stirring the tank to ensure larvae are uniformly distributed, followed by extraction of six 5 mL samples. Each sample is then manually counted under a stereo microscope, which takes on average 20 minutes per tank sampled. Ideally, sampling would be performed hourly over the seven day larval rearing period, to enable implementation of mitigating strategies during rapid culture declines. However, the manual effort required typically limits this to once daily per tank.  The repeated tank stirring can also have a negative impact on the coral spawn, and therefore culture health~\cite{randall2020rapid}. 

\subsection{Automated Coral Spawn Monitoring}

Previous works on automated spawn counting have been limited by cost and functional usability. Randall~\etal used large particle flow cytometry\footnote{Cytometry is a laboratory technique that measures the number and characteristics of cells.} to count coral spawn; however, the required flow cytometer was prohibitively expensive (\ie $\approx$500,000 USD) and the methods were not easily scalable or integrated with aquaculture facility tanks~\cite{randall2020rapid}.  Mullen~\etal developed an underwater microscope with a microscopic objective in front of an electrically tunable lens for long-term benthic ecosystem studies~\cite{mullen2016underwater}, which was later improved in~\cite{ben2025benthic}, with the incorporation of focal scans and estimates of photosynthetic efficiency.  Shahani~\etal used a variable objective lens in their underwater benthic microscope to achieve a higher depth of field imagery via focus stacking~\cite{shahani2021design}. While these solutions~\cite{mullen2016underwater, ben2025benthic, shahani2021design} demonstrate advances in underwater microscopy, they are not suitable for automated spawn monitoring or designed for integration in large-scale aquaculture facilities.

Several commercially available scientific and industrial imaging products exist for use in aquatic environments (\eg for monitoring plankton biodiversity)~\cite{cpics, pi10, ISIIS-DPI}. Recent advances include compact integrated cameras with embedded computer chips (\eg Luxonis, Arducam, Point Grey).  Some of these commercial sensors have been used in the field at macro scales for coral reef conservation projects, including for mapping reefs~\cite{dunbabin2020uncrewed} and distributing coral larvae across reefs~\cite{mou2022reconfigurablerobots}. While potentially useful for field applications, the optics used in these systems are unsuitable for coral conservation aquaculture workflows, due to the microscopic scales of coral spawn imaging. Additionally, most commercially-available housings lack the robustness needed to operate in saltwater environments.

\subsection{Deep Learning for Scientific Counting Applications}

Deep learning and computer vision has increased the capabilities of counting-based applications in many scientific monitoring applications, including for cell counting in health applications~\cite{grishagin2015automatic, morelli2021automating, falk2019u}. Adoption of these technologies has led to ten times faster operations and yields more reliable and consistent results compared to manual counting~\cite{grishagin2015automatic}.

In marine science and aquaculture, deep learning has been successfully used for counting plankton eggs~\cite{weldrick2022promising}, shrimp eggs~\cite{zhang2021shrimp}, and fish~\cite{ditria2021annotated, connolly2021improved}, however, it has not yet been used for coral spawn counting. 

Semi-automated human-in-the-loop annotation reduces the burden of fully manual labeling by combining model predictions with iterative human validation~\cite{garcia2023ten, dayoub2017episode, desmond2021semi}, which has been applied in ecology, including labeling large camera trap datasets and segmenting coral reef imagery~\cite{norouzzadeh2021deep, bakana2023digital, raine2024human}. Here, we adopt human-in-the-loop active learning to efficiently annotate coral spawn imagery collected by CSLICS.




\section{Method}
\label{sec:method}

We propose the Coral Spawn and Larvae Imaging Camera System (CSLICS), a submersible computerized camera to streamline counting during the embryonic and larval rearing phases of coral aquaculture workflows.  We combine commercially-available, low-cost and modular components with state-of-the-art image processing techniques to automatically monitor coral spawn. The following sections detail the coral aquaculture process using CSLICS for automated spawn monitoring (Section~\ref{subsec:process}); the hardware design (Section~\ref{subsec:hardware}), including the system requirements (Section~\ref{subsubsec:requirements}); architecture (Section~\ref{subsubsec:architecture}); and key components (Section~\ref{subsubsec:components}).  The software design is explained in Section~\ref{subsec:software}, and includes a description of the coral detection and counting models used.

\subsection{Coral Aquaculture with CSLICS} 
\label{subsec:process}
To automate and increase the frequency of coral counting, we propose the following CSLICS process and timings during aquaculture monitoring of early coral life stages for species belonging to the family \textit{Acroporidae} (see Fig.~\ref{fig:cslicsmodes}). The process is outlined in the context of using 500 L conical fiberglass larval culture tanks supplied with flow-through ({$\approx$}125 L/h) filtered seawater (0.04 $\mu$m, typically 26-29°C)~\cite{pollock2017coral, severati2024autospawner}.
\begin{itemize}
    \item t\textsubscript{0}: The tanks are running at a known volume and are being stocked at 1 cell/mL with coral spawn, consisting of fertilized and unfertilized eggs from the preceding aquaculture stages.
    \item t\textsubscript{0}--t\textsubscript{1}: The coral spawn is undergoing embryogenesis, leading to a cleavage of the cells through multiple cell development stages until ciliated larvae\footnote{Larvae that have cilia, or tiny hair-like projections, that help them move. They are found in many marine invertebrates.} are formed. During the first {$\approx$}12 hours of development, the cells are positively buoyant and  extremely fragile. The aeration of the tank is therefore kept at a minimum ({$\approx$}10 L/h) to minimize water surface disturbance, while water inlets facilitate a gentle surface movement for optimized CSLICS analysis without affecting the spawn. CSLICS images the surface of the water (Fig.~\ref{fig:cslicsmodes}) and analyzes the fertilization success.
    \item t\textsubscript{1}: {$\approx$}12-24 hours following fertilization, the embryogenesis has advanced enough to not be impacted by excessive water movement, \ie, prawn-chip stage\footnote{A stage in the development of embryos, when they assume a concave-convex dish shape.}, allowing an increase in the aeration of the tank to facilitate homogenization throughout the water column ({$\approx$}100 L/h). Consequently, the water surface becomes disturbed from the air bubbles making continued surface operation of CSLICS unsuitable. CSLICS is switched to sub-surface mode (Fig.~\ref{fig:cslicsmodes}) to analyze the developing larvae in the water column.
    \item t\textsubscript{1}--t\textsubscript{2}: The coral larvae continue to develop until they are mature enough to successfully undergo metamorphosis into a coral recruit. 
    \item t\textsubscript{2}:  After approximately six days, the coral larvae are typically mature enough to undergo settlement and are therefore ready for harvesting from the larval culture tank. The final CSLICS density counts are used to calculate the proportion of the culture tank to harvest, based on the number of available settlement substrate units and the target larval density required in each to achieve the desired settlement numbers (\eg~50 larvae/L of the settlement tank).
\end{itemize}

\subsection{Hardware Design}
\label{subsec:hardware}

CSLICS is designed for use in coral aquaculture facilities, typically in combination with larval culture tanks similar to that described in Section~\ref{subsec:process}, which presents several challenges in terms of operational environment, system performance and design.

This section outlines the key system requirements which informed our system architecture and hardware design.

\subsubsection{System Requirements}
\label{subsubsec:requirements}

The system requirements are informed by coral production workflow objectives and constraints imposed by coral biology and the operational environment.  The CSLICS system must:
\begin{enumerate}[label=\alph*)]
    \item measure the fertilization success ratio (defined in Section~\ref{subsec:evalmetrics}, equation~\ref{eq:fert}) for the first two hours following tank stocking to provide critical data used to assess the coral larval culture quality;
    \item provide the total coral spawn counts of the tank at least once per hour for the duration of the larval rearing period (up to seven continuous days) to help operators observe trends in the larval rearing process;
    \item image early coral life stages ranging from 150-500 $\mu$m in diameter in the relatively low-light operating conditions of the aquaculture facility;
    \item not contain any exposed metal alloys that are toxic to the corals (\eg copper);
    \item be waterproof (\eg at least IP68) to avoid water damage from possible submersion, reduce electrical hazard, and minimize corrosion from the saltwater environment of the aquaculture facility;
    \item not have 240 V components closer than 1 m from the water (in accordance with safety regulations); and
    \item minimize physical interaction with the coral spawn to reduce interruption to the rearing tank's water flow and potential risks to the corals.
\end{enumerate}

\subsubsection{System Architecture}
\label{subsubsec:architecture}
\begin{figure}[t]
    \centering
    \includegraphics[width=0.95\columnwidth,clip,trim=2.3cm 2.5cm 1.1cm 0cm]{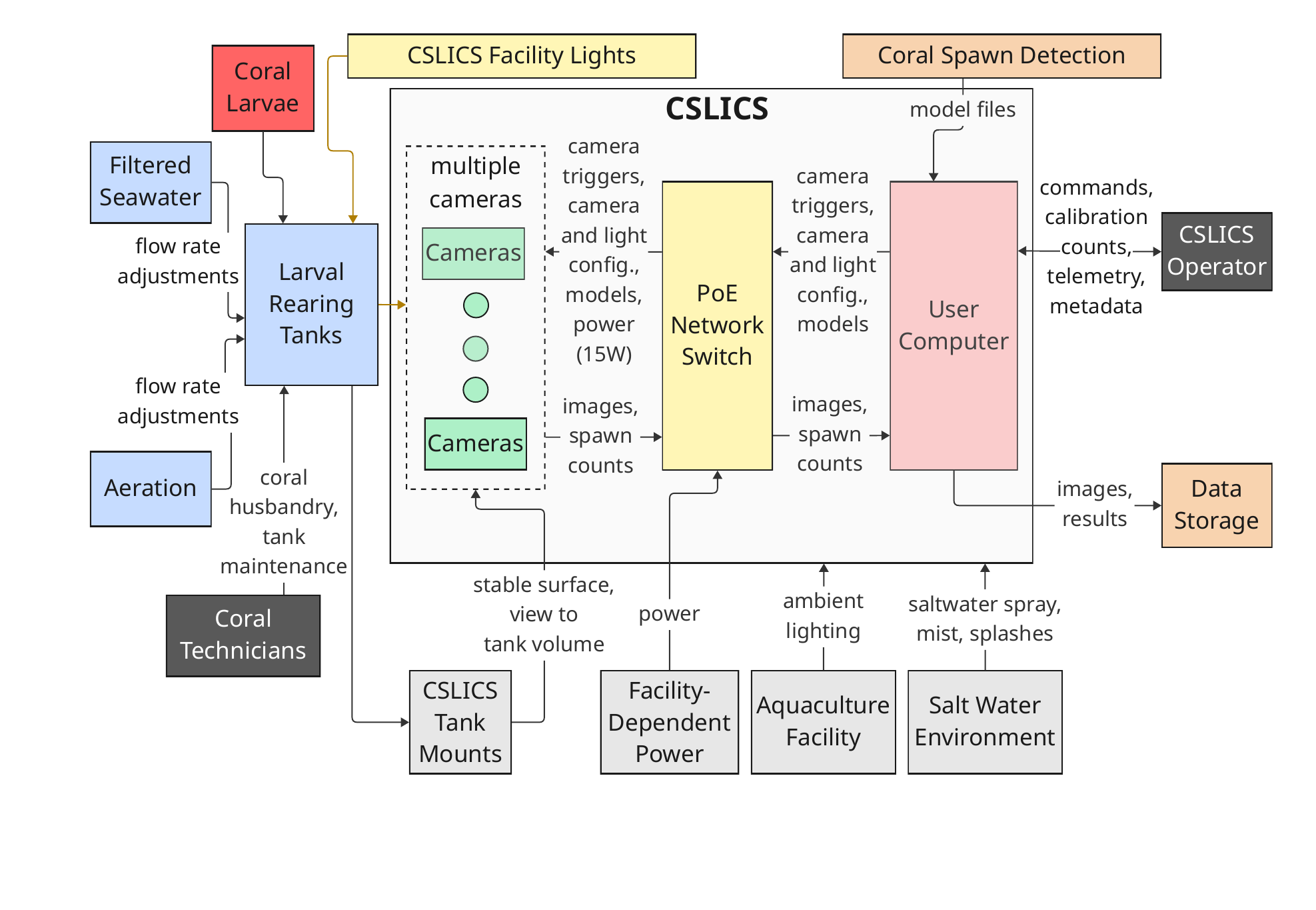}
    \vspace*{-0.4cm}
    \caption{The CSLICS system boundary diagram, illustrating the major inputs and outputs.}
    \label{fig:boundary}
    \vspace{-0.5cm}
\end{figure}

CSLICS is an image capture and analysis system comprising of a camera and controller mounted to the tank~(Fig.~\ref{fig:boundary}). The system connects via Ethernet to a monitoring computer that manages user commands, telemetry from the image capture system, and handles the captured data. Decentralized image processing (\ie onboard computation) was adopted to facilitate system scalability, reduce bandwidth requirements, and minimize the computational load on the central user computer, particularly when operating with 60 or more camera units.

\subsubsection{Components}
\label{subsubsec:components}

As shown in Fig.~\ref{fig:cslics}, the key components are a Raspberry Pi High Quality Camera with a Sony IMX477 sensor and a Pimoroni microscope lens at 0.12-1.8x magnification, which provides sufficient spatial resolution of coral spawn at typical working distances of 5 cm from the water surface.  A Raspberry Pi 4 Model B (8 GB) controls the cameras, responds to the central computer commands and saves the images during data collection.  Watertight enclosures from Blue Robotics are used to house the camera and computer hardware.  A Power-over-Ethernet (PoE) board delivers power and communications to CSLICS.  The network is managed by an 8-port PoE switch, delivering both power and communications via a single cable. The PoE connection provides 15 W of power at 48 V, sufficient for running the onboard computer, camera and storage device.  The managed switch enables remote and scheduled power cycling of the CSLICS without having to access the network switch itself, and allows multiple CSLICS modules to be connected together. CSLICS is mounted to the tank using modular photography clamps.  The cost of a single CSLICS camera unit is $\approx$1,000 USD. 




\subsection{Software Design}
\label{subsec:software}


When powered on, a camera bring-up service starts the CSLICS system, which configures and triggers the camera via the Robot Operating System (ROS)~\cite{quigley2009ros}. Captured images are saved with metadata and then processed by the appropriate surface or sub-surface coral detection model.

We train two YOLOv8 object detectors~\cite{redmon2016you, ultralyticsyolov8}, which were selected for their accessibility, ease-of-use, widely supported framework, extensive documentation, and the variety of model sizes available.  The YOLOv8x 640p model was chosen as a suitable balance between inference speed and accuracy, enabling detection on the CSLICS hardware at 2~seconds per frame. We train one model for surface detection of coral spawn, which is capable of detecting coral spawn at different stages of embryogenesis (Fig.~\ref{fig:subsurface} and Section~\ref{subsec:dataformat}).  We also train a model for sub-surface detection of the corals, where it is no longer necessary to differentiate between developmental stages, so this is a single-class coral detector.  

We leverage a semi-supervised, human-in-the-loop labeling approach to reduce the annotation effort required and to rapidly label a dataset for training the detection models~\cite{mosqueira2023human}.  We manually annotate a small number of images to train a preliminary object detection model, which is iteratively updated with additional samples obtained by reviewing predictions from the model. Further details on the datasets and labeling approaches are available in Section~\ref{subsec:dataformat}.


\section{Experimental Setup}
\label{sec:experimentalsetup}

\subsection{Data Collection}
\label{subsec:datacollection}

Data was collected at a coral aquaculture facility during two mass spawning events in 2022 and 2023. The selection of coral species was primarily determined by marine scientists based on factors such as the availability of gravid corals\footnote{Corals that are carrying eggs and are ready for spawning.}, quality of coral spawn, number of parental colonies, and potential for long-term deployment. The primary focus was on commonly occurring and fast growing branching corals, with two species from the family \textit{Acroporidae} initially targeted (\textit{Acropora kenti} and \textit{Acropora loripes}). Additionally, data for species belonging to other coral families (\eg \textit{Mycedium elephantotus} and \textit{Dipsastraea favus}) were collected for future research and expansion of the CSLICS detector models.

The larval culture tanks (Section~\ref{subsec:process}) were filled to the appropriate water level approximately two hours prior to the expected spawning time of the target species. At this time, CSLICS units were mounted and focused at the water level for surface-based imaging. CSLICS were mounted 5 cm above the water surface, depending on the lens position within the housing. Three CSLICS units were mounted per larval culture tank.

\begin{figure}[t]
    \centering
    \includegraphics[width=0.47\columnwidth,clip,trim=3cm 2.61cm 5cm 2cm]{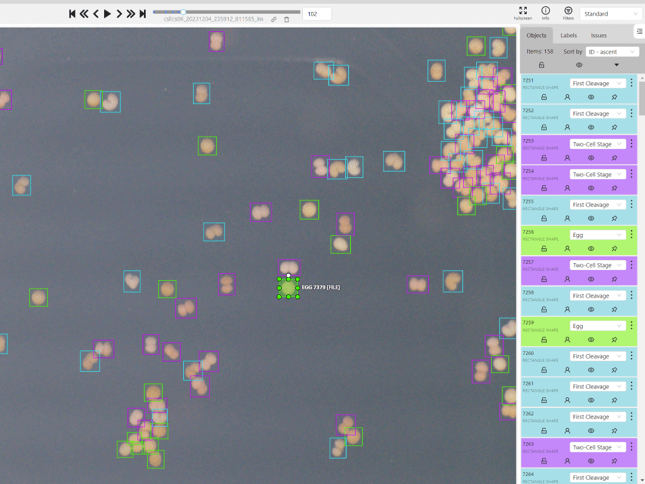}
    \hfill
    \includegraphics[width=0.47\columnwidth,clip,trim=1cm 0cm 1cm 0.45cm]{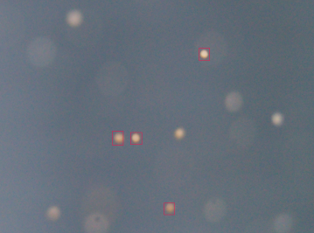}
    \caption{Manual annotation of CSLICS images: surface (left) and sub-surface (right), highlighting the challenging visual characteristics of the coral spawn.  The sub-surface detection task (right) is further complicated as we aim to detect only corals that are in-focus.}
    \label{fig:subsurface}
    \vspace{-0.4cm}
\end{figure}

Once the culture tank was stocked with coral spawn, CSLICS were activated to capture images continuously at a rate of one image every 10 seconds, to allow for image capture and neural network processing time. This resulted in approximately 360 images per hour, allowing for comprehensive sampling of the corals over time. It was important to collect image data over a sufficiently large period of time in order to average inhomogeneity caused by coral slicks.  Manual counts were taken immediately after stocking the culture tanks, and then once daily for comparison to the CSLICS method, as detailed in Section~\ref{subsec:process}. The operators occasionally adjusted CSLICS focus due to fluctuating water levels resulting from flow rate or filtration adjustments. 

Approximately 24 hours after tank stocking, the operators increased the tank aeration to encourage homogeneity of coral density throughout the tank (Fig.~\ref{fig:cslicsmodes}). At this point, CSLICS were lowered to submerge the units' front element no more than 1 cm into the water, to minimize interaction with the coral spawn, marking the beginning of the sub-surface operation. The exact time was recorded by the operators, and CSLICS were then left to continuously capture and process data until the completion of the larval rearing process (6.5 days).


\begin{table*}[t]
\setlength{\tabcolsep}{2px}
\vspace*{0.15cm}
\caption{Surface Coral Detection Results, where P is Precision, R is Recall, F1 is F1 Score, AP is the Average Precision \texttt{@}0.5 and mAP is the mean Average Precision \texttt{@}0.5 (Refer to Section~\ref{subsec:evalmetrics} for Metric Definitions)} 
\vspace*{-0.15cm}
\label{tab:surface}
\centering
\resizebox{\textwidth}{!}{%
\begin{tabular}{@{} lccccccc @{}} 
 \toprule
\multirow{2}[3]{*}{\textbf{Method}} 
& \textbf{Egg} 
& \textbf{First Cleavage} 
& \textbf{Two-Cell} 
& \textbf{Four-to-Eight Cell} 
& \textbf{Advanced} 
& \textbf{Damaged} 
& \textbf{Overall} \\
 \cmidrule(lr){2-2} \cmidrule(lr){3-3} \cmidrule(lr){4-4} \cmidrule(lr){5-5} \cmidrule(lr){6-6} \cmidrule(lr){7-7} \cmidrule(lr){8-8}
& P / R / F1 / AP & P / R / F1 / AP & P / R / F1 / AP & P / R / F1 / AP & P / R / F1 / AP & P / R / F1 / AP & F1 / F1 ex. Dam. / mAP \\
\midrule
YOLOv8~\cite{ultralyticsyolov8} 
& 91.1 / 89.9 / 90.5 / 90.7
& 86.5 / 76.0 / 80.9 / 83.5
& 82.5 / 84.8 / 83.6 / 87.1
& 75.6 / 70.2 / 72.8 / 74.1
& 80.2 / 88.3 / 84.1 / 87.5
& 66.2 / 37.8 / 48.1 / 52.1
& 76.7 / 82.4 / 79.2 \\
\arrayrulecolor{black!100}\bottomrule 
\end{tabular}
}
\vspace*{-0.25cm}
\end{table*}

 

\subsection{Dataset Annotation and Format}
\label{subsec:dataformat}

We collected and labeled two datasets, one for surface mode operation and one for sub-surface mode operation. 

\subsubsection{Surface Data}

Over the 2022 and 2023 spawning events, 2000 images were annotated with bounding boxes using the human-in-the-loop approach described in Section~\ref{subsec:software}.  We first annotated 100 images manually to train a preliminary object detection model, which was used to pseudo-label the next 200 images. These predictions were then manually reviewed before re-training the model. This process was repeated until all 2000 images were labeled.  The labeled images were randomly partitioned into 70\%/20\%/10\% splits for training, validation and testing subsets, respectively.

The target classes for surface-based imaging are shown in Fig.~\ref{fig:cslics}: egg, first cleavage, two-cell stage, four-to-eight cell stage, advanced and damaged. The egg class is typically circular. It is visually challenging to determine if the egg is fertilized; therefore, we wait for the cell to start cleaving to indicate fertilization. First cleavage is when the egg starts to cleave into two cells, forming a pinch on one side, similar to a peach. The two-cell stage is when the coral is forming a pinch on two opposing sides clearly showing the definition of two cells. The four-to-eight cell stage is when the coral is composed of four to eight cells, denoted due to its potential ambiguity of having four pairs of cells stacked (the eight-cell stage can appear like a four-cell stage).  The advanced stage is a class that encompasses any further embryo development, from 16-cells to prawn-chip and beyond. For the purposes of CSLICS, further classification of embryo development is not required. Finally, there are damaged cells, which appear as elongated or misshapen eggs or embryos, and typically occur due to inadequate fertilization or over-handling.

\subsubsection{Sub-surface Data}
When the corals are dispersed in the water column, the limited depth of field of the optics becomes strongly apparent (\eg Fig.~\ref{fig:subsurface}).  Corals too close or too far from the camera appear out-of-focus, while those in-focus remain sharp and defined. This is why we perform single-class coral detection for sub-surface imaging.  The exact embryogenesis stage is also less critical at this point as the developing corals approach their final developmental stage (planula larvae) with a smooth and oval appearance. Importantly, the sub-surface detector only counts corals that are in focus, and we therefore manually label this data using CVAT (\url{https://www.cvat.ai/}), and ensure that only in-focus corals are annotated.

\subsection{Evaluation Metrics}
\label{subsec:evalmetrics}
We report the performance of CSLICS in terms of surface detection, fertilization success, sub-surface detection, and sub-surface tank counts.

\subsubsection{Surface Detection}
We use precision (P) and recall (R) to quantify the performance of the model's surface detections. Precision is the ratio of correctly predicted positive detections to the total predicted positives, and recall is the ratio of correctly predicted positive detections to all the observations in the actual class. We balance these metrics by considering the F1 score, which is the harmonic mean of precision and recall.  We also calculate the common object detection metric Average Precision (AP), which is the area under the precision-recall curve. When calculating these metrics, we use a confidence threshold of 0.5, and an IoU threshold of 0.5. We calculate the precision, recall, F1 score and AP\texttt{@}0.5 for each class, as well as finding a macro F1 score and mAP\texttt{@}0.5 to indicate the overall detector performance.

\subsubsection{Fertilization Success}
Fertilization success is a ratio of fertilized eggs to total viable eggs (fertilized and unfertilized). It can be measured by comparing the ratio of classes from the coral detector model. Let $n_e$, $n_c$, $n_2$, $n_4$, $n_a$ represent the number of eggs, first cleavage, two-cell, four-to-eight-cell, and advanced spawn counts, respectively. Note, damaged spawn counts are not included as they are considered to be nonviable.  Then the fertilization success ratio $f$ is given as,
\begin{equation}
    f = \frac{n_c + n_2 + n_4 + n_a}{n_e + n_c + n_2 + n_4 + n_a}.
    \label{eq:fert}
\end{equation}
This is a convenient metric because it is bounded from 0 to 1, with 0 implying either a culture containing only eggs, and~1 being a completely fertilized culture.
Thus, the fertilization success can be calculated on a per-image basis.



\subsubsection{Sub-surface Detection}
\label{subsubsec:subsurface_detection}
As coral spawn mature, they transition from the surface to water column, partially due to developing cilia that help them swim, but largely due to the tank's hydrodynamics once the filtration rate is increased (Fig.~\ref{fig:cslicsmodes}).  To obtain detection counts when CSLICS is in sub-surface operation, we used the sub-surface detection model to produce per-image counts of the recruits that appear in focus. This is a binary detection case, which we evaluate using precision, recall, F1 score and the mAP\texttt{@}0.5. 

\subsubsection{Sub-surface Tank Counts}
\label{subsubsec:subsurface_counts}
We convert the image-based counts into tank counts using a scaling factor.  As this scaling factor is dependent on many variables, such as the initial manual spawn input, the tank volume, the coral species and the target density, we use the first instance of manual counts at the start of the sub-surface segment to determine a scaling factor to convert the image-based counts to tank counts.

\begin{figure*}[t]
    \centering
    \includegraphics[width=0.8\columnwidth,clip,trim=0.6cm 0.6cm 0.6cm 0.5cm]{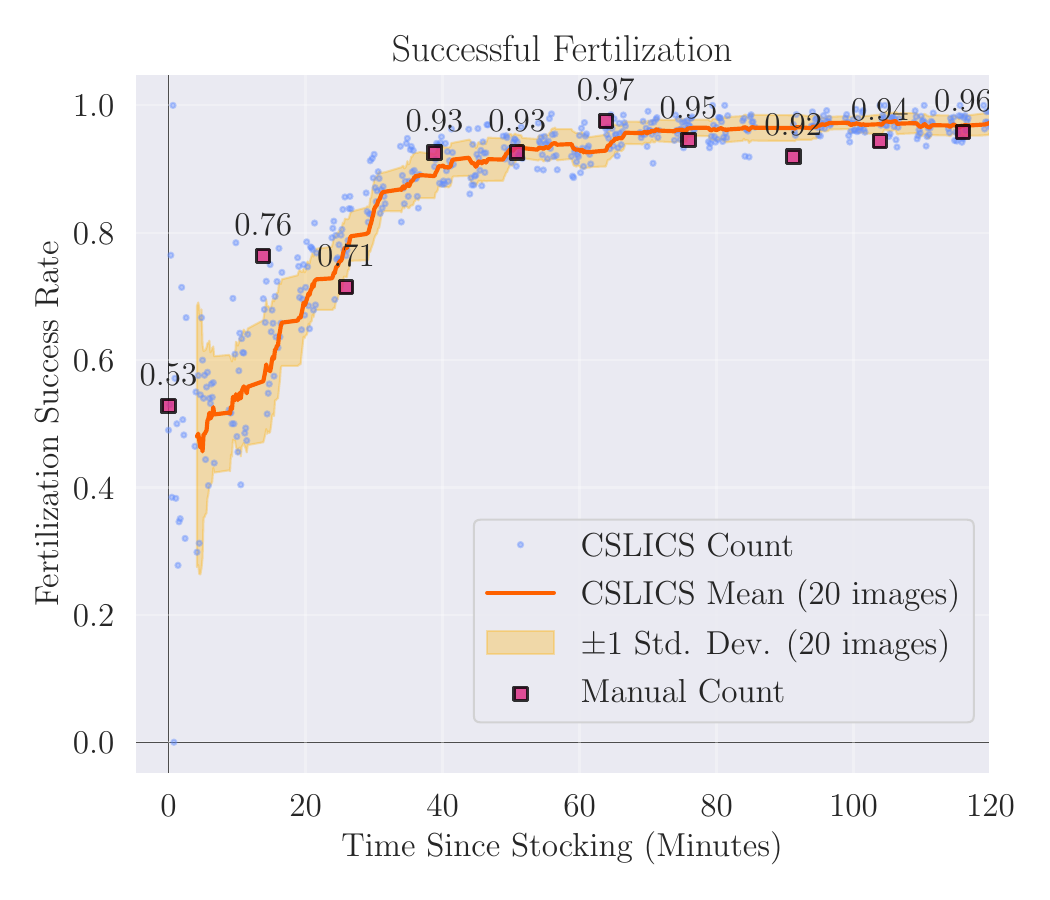}
    \hspace{0.5cm}
    \includegraphics[width=0.8\columnwidth,clip,trim=0.6cm 0.6cm 0.6cm 0.5cm]{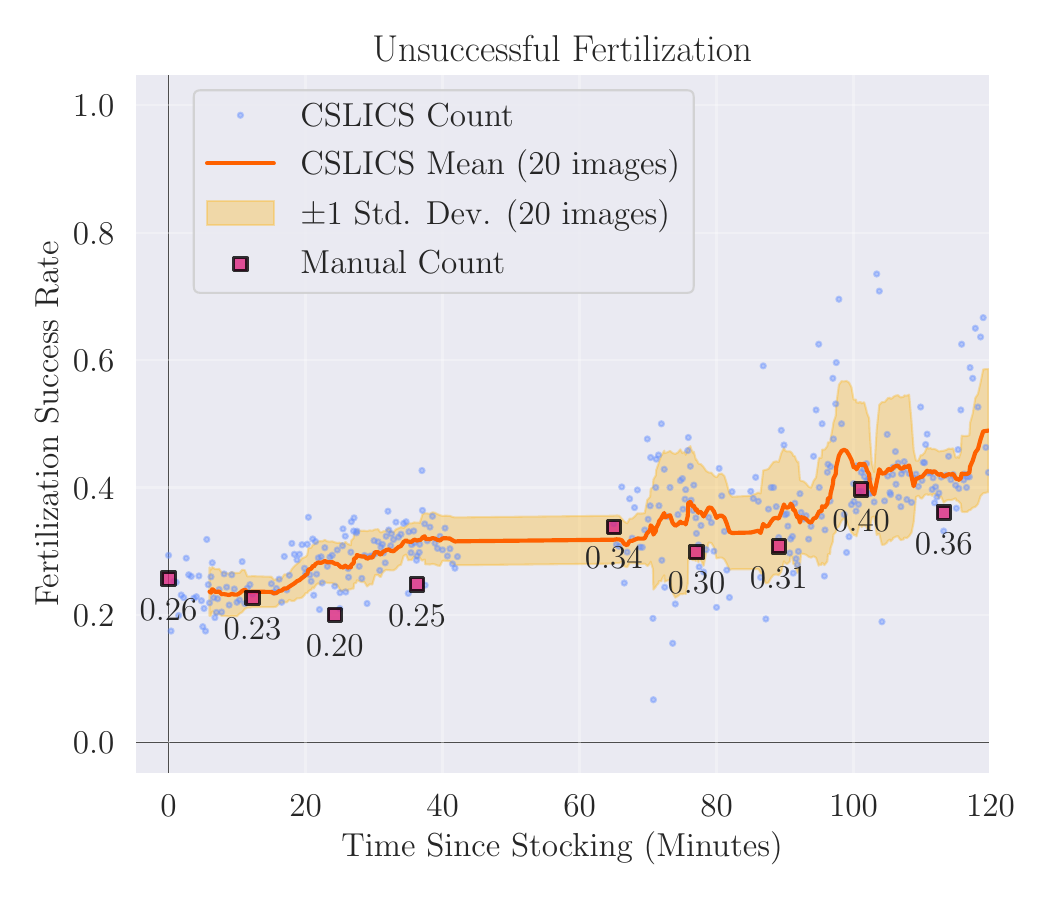}
    \caption{Fertilization rate over time for two tanks. Left: a successful larval culture with a Root Mean Square Error (RMSE) of 0.101. Right: an unsuccessful larval culture with RMSE 0.131. CSLICS is able to clearly identify fertilization success and provide early warning of deteriorating cultures.}
    \vspace{-0.4cm}
    \label{fig:fert}
\end{figure*}

\section{Results}
\label{sec:results}

We evaluate our CSLICS approach for monitoring coral spawn in aquaculture for both the surface and sub-surface operational modes. When in surface mode, we report the surface detections and the fertilization success (Section~\ref{subsec:results-surface}).  For sub-surface operation, we report the sub-surface detections and the total tank counts (Section~\ref{subsec:results-subsurface}). For both fertilization success and total spawn counts, we compare against human-derived estimates. Overall, we find that CSLICS effectively monitors rates of fertilization and can clearly differentiate between successful and unsuccessful larvae cultures.  

Further, we find that CSLICS provides a time saving of 5,720 hours per spawning event. This value is derived by calculating the effort required for equivalent manual sampling: 12 samples per hour for the surface operation (12 hours/144 samples), plus one sample per hour for sub-surface operation (six days/144 samples), resulting in 288 samples per tank. For 60 tanks and 20 minutes per sample, manual sampling would require 5,760 hours of labor, minus 40 hours for one person to operate the CSLICS system.

\subsection{Surface Operation}
\label{subsec:results-surface}

\subsubsection{Coral Development Stage Detection}

We report the performance of our multi-class coral detector in Table~\ref{tab:surface}. Our detector achieves greater than 72.8\% F1 score for all development classes, and 48.1\% for the damaged class. This class suffers in performance due to the relatively low number of examples (839 instances, compared with 1,700-5,300 instances for the other classes) in the training set, as well as the challenging and highly variable visual characteristics of damaged corals.  As the damaged class is not included in the fertilization success ratio calculations, we also report the overall F1 score excluding this class, which is 82.4\%. 

\begin{table}[t]
\setlength{\tabcolsep}{3px}
\vspace*{0.15cm}
\caption{Sub-surface Coral Detection Results (Refer to Section~\ref{subsec:evalmetrics} for Metric Definitions)} 
\vspace*{-0.2cm}
\label{tab:subsurface}
\centering
\scriptsize
\begin{tabularx}{\columnwidth}{@{}l>{\centering\arraybackslash}X>{\centering\arraybackslash}X>{\centering\arraybackslash}X>{\centering\arraybackslash}X>{\centering\arraybackslash}X}
\toprule
\textbf{Method} & \textbf{Precision} & \textbf{Recall} & \textbf{F1 Score} & \textbf{mAP\texttt{@}0.5}\\
\midrule
YOLOv8~\cite{ultralyticsyolov8} & 79.0 & 87.8 & 83.0 & 87.9 \\
\arrayrulecolor{black!100}\bottomrule 
\end{tabularx}
\vspace*{-0.3cm}
\end{table}

\subsubsection{Fertilization Success}

CSLICS was effectively used to image the surface of the larval rearing tanks and detect corals of different embryonic development stages.  From these detections, Eq.~\ref{eq:fert} was used to determine the fertilization success. Fig.~\ref{fig:fert} shows the fertilization rate over time for two tanks, each equipped with CSLICS.

A rolling mean for the fertilization success rate is shown in orange, where the mean is calculated with a window of 20 images (thus averaged over $\approx$3 minutes). This figure demonstrates CSLICS capability to predict a successful fertilization for a tank: the left plot shows the curve for a successful fertilization, as the rate of fertilized to non-fertilized eggs increases steadily over time.  The right plot instead shows an unsuccessful larval culture, as indicated by a fertilization rate which remains significantly lower than the successful culture for the duration (Fig.~\ref{fig:fert}).  This occurs as unfertilized eggs begin to dissolve and cause a chain reaction throughout the larval culture.




\subsection{Sub-surface Operation}
\label{subsec:results-subsurface}

\subsubsection{Coral Detection}
We report the precision, recall, F1 score and mAP\texttt{@}0.5 of our coral sub-surface detector model in Table~\ref{tab:subsurface}.  These results demonstrate that we are able to detect and count coral spawn in the sub-surface setting, with an F1 score of 83.0\% and 87.9\% mAP\texttt{@}0.5.  The sub-surface setting is more challenging than the surface case (as seen in Fig.~\ref{fig:subsurface}).  As the corals reach the settlement stage (Fig.~\ref{fig:cslicsmodes}), they begin to descend in the tank, and therefore become out-of-focus of the camera.  Further, as the camera housing is submerged, it can become coated with coral larvae residue.  This limitation will be addressed in the next iteration of the system design, which will feature a camera imaging the tank through a viewing window on the side of the tank.

\subsubsection{Sub-surface Total Tank Counts}

\begin{figure}[t]
    \centering
    \includegraphics[width=0.85\columnwidth,clip,trim=0.6cm 0.6cm 0.6cm 0.5cm]{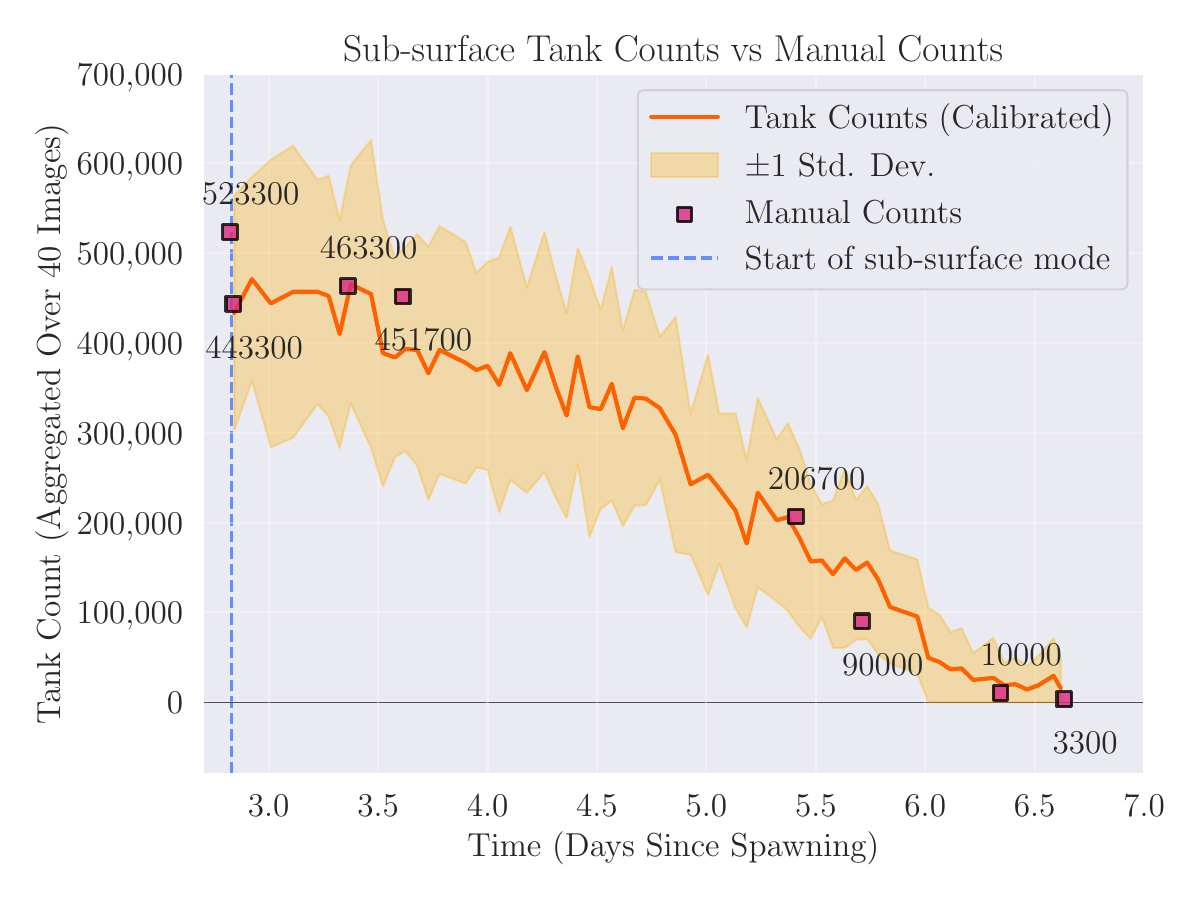} \\
    \vspace{-0.1cm}
    \caption{CSLICS sub-surface tank counts, demonstrating close alignment between CSLICS estimates and manual counts (RMSE = 45,670 corals), sufficient for monitoring purposes.}
    \label{fig:subsurfacetank}
    \vspace{-0.3cm}
\end{figure}

In sub-surface operation we quantify the spawn counts for the entire tank. The process for estimating total spawn count was described in Section~\ref{subsubsec:subsurface_counts}. 
Fig.~\ref{fig:subsurfacetank} shows CSLICS sub-surface total tank counts for a culture in December 2023. The rolling mean tank count is shown with a centered window of 40 images, with one standard deviation bounds shown in the lighter colors, and the manual counts are shown as the square markers. To the best of our knowledge, the manual counts were consistent as they were sampled and counted by the same personnel following the same protocol. This figure demonstrates the accuracy of CSLICS tank count estimates. 


\section{Conclusion}
\label{sec:conclusion}

In this work, we proposed an automated coral spawn and larvae imaging camera system to enhance high-density larval rearing in coral aquaculture.  We presented the two operational modes of the system (surface and sub-surface), including the system architecture, hardware specifications, software implementation, and the data collection and annotation methods. We trained two models for counting corals of species belonging to the family \textit{Acroporidae}: one tailored to detecting coral spawn on the surface across different developmental stages, and the other designed for sub-surface spawn detection. The data captured by CSLICS illustrated clear trends between successful and unsuccessful fertilization success ratios to assess the viability of the coral spawn cultures. Additionally, CSLICS provided accurate tank counts for the sub-surface operation, showing comparable trends with manual counts. 

CSLICS' automation and vision-based approach reduced coral disturbance and enabled recording frequent observations into a digital database. This data supports correlation analysis of treatment strategies with larval rearing outcomes. CSLICS enables robust and consistent tracking of coral spawn at different stages of development.  Further, CSLICS streamlines coral rearing processes and avoids the 5,720 hours of labor per spawning event that would be required for manual sampling at the same frequency.

The next iteration of CSLICS is currently in development, which will improve detection of damaged cells, and feature side-mounted imaging to reduce the impacts of coral larvae residue build-up on the housing during sub-surface operation.


\vspace*{-0.1cm}
\bibliographystyle{IEEEtran}
\bibliography{bibliography}

@IEEEtranBSTCTL{bstctl:forced_etal,
  CTLuse_forced_etal = "yes",
  CTLmax_names_forced_etal = "6",
}

@IEEEtranBSTCTL{bstctl:nodash,
  CTLdash_repeated_names = "no",
}

@inproceedings{quigley2009ros,
  title={{ROS: A}n open-source {Robot Operating System}},
  author={Quigley, Morgan and Conley, Ken and Gerkey, Brian and Faust, Josh and Foote, Tully and Leibs, Jeremy and Wheeler, Rob and Ng, Andrew Y and others},
  booktitle={IEEE Conference on Robotics and Automation Workshop on Open Source Software},
  year={2009},
}

@article{costanza2014changes,
  title={Changes in the global value of ecosystem services},
  author={Costanza, Robert and De Groot, Rudolf and Sutton, Paul and Van der Ploeg, Sander and Anderson, Sharolyn J and Kubiszewski, Ida and Farber, Stephen and Turner, R Kerry},
  journal={Global Environmental Change},
  volume={26},
  pages={152--158},
  year={2014},
  publisher={Elsevier}
}

@article{heron2017impacts,
  title={Impacts of climate change on World Heritage coral reefs: A first global scientific assessment},
  author={Heron, Scott Fraser and Eakin, Carlon Mark and Douvere, Fanny and Anderson, Kristen L and Day, Jon C and Geiger, Erick and Hoegh-Guldberg, Ove and Van Hooidonk, Ruven and Hughes, Terry and Marshall, Paul and others},
  year={2017},
  journal={UNESCO}
}

@article{de201227,
  title={The 27--year decline of coral cover on the {Great Barrier Reef} and its causes},
  author={De’Ath, Glenn and Fabricius, Katharina E and Sweatman, Hugh and Puotinen, Marji},
  journal={Proceedings of the National Academy of Sciences},
  volume={109},
  number={44},
  pages={17995--17999},
  year={2012},
  publisher={National Acad Sciences}
}

@misc{ultralyticsyolov8,
  title = {Ultralytics {YOLOv8} Github Repository},
  author={Glenn Jocher},
  howpublished = {\url{https://github.com/ultralytics/ultralytics}},
  note = {Accessed: 2023-02-04}
}

@inproceedings{redmon2016you,
  title={{You Only Look Once: U}nified, real-time object detection},
  author={Redmon, Joseph and Divvala, Santosh and Girshick, Ross and Farhadi, Ali},
  booktitle={IEEE Conference on Computer Vision and Pattern Recognition},
  pages={779--788},
  year={2016}
}

@article{mosqueira2023human,
  title={Human-in-the-loop machine learning: A state of the art},
  author={Mosqueira-Rey, Eduardo and Hern{\'a}ndez-Pereira, Elena and Alonso-R{\'\i}os, David and Bobes-Bascar{\'a}n, Jos{\'e} and Fern{\'a}ndez-Leal, {\'A}ngel},
  journal={Artificial Intelligence Review},
  volume={56},
  number={4},
  pages={3005--3054},
  year={2023},
  publisher={Springer}
}

@article{grishagin2015automatic,
  title={Automatic cell counting with {ImageJ}},
  author={Grishagin, Ivan V},
  journal={Analytical Biochemistry},
  volume={473},
  pages={63--65},
  year={2015},
  publisher={Elsevier}
}

@inproceedings{raine2024human,
  title={Human-in-the-loop segmentation of multi-species coral imagery},
  author={Raine, Scarlett and Marchant, Ross and Kusy, Brano and Maire, Frederic and Sunderhauf, Niko and Fischer, Tobias},
  booktitle={IEEE/CVF Conference on Computer Vision and Pattern Recognition},
  pages={2723--2732},
  year={2024}
}

@INPROCEEDINGS{dunbabin2020uncrewed,
  author={Dunbabin, Matthew and Manley, Justin and Harrison, Peter L.},
  booktitle={OCEANS}, 
  title={Uncrewed Maritime Systems for Coral Reef Conservation}, 
  year={2020},
  doi={10.1109/IEEECONF38699.2020.9389173}}

@article{randall2020rapid,
  title={Rapid counting and spectral sorting of live coral larvae using large-particle flow cytometry},
  author={Randall, Carly J and Speaks, Justin E and Lager, Claire and Hagedorn, Mary and Llewellyn, Lyndon and Pulak, Rock and Thompson, Julia and Bay, Line K and Mead, David and Heyward, Andrew J and others},
  journal={Scientific Reports},
  volume={10},
  number={1},
  pages={12919},
  year={2020},
  publisher={Nature Publishing Group UK London}
}

@misc{cpics,
  title = {{CPICS - Continuous Particle Imaging Classification System}},
  howpublished = {\url{https://www.coastaloceanvision.com/cpics}},
  note = {Accessed: 2024-05-01}
}

@misc{pi10,
  title = {{Plankton Analytics - An Overview of the Pi-10}},
  howpublished = {\url{https://www.planktonanalytics.com/}},
  note = {Accessed: 2025-09-10}
}

@misc{ISIIS-DPI,
  title = {{In-Situ Ichthyoplankton Imaging System-Deep-Focus Particle Imager (ISIIS-DPI)}},
  howpublished = {\url{https://www.isiis-dpi.com/}},
  note = {Accessed: 2025-09-10}
}

@article{mullen2016underwater,
  title={Underwater microscopy for \emph{in situ} studies of benthic ecosystems},
  author={Mullen, Andrew D and Treibitz, Tali and Roberts, Paul LD and Kelly, Emily LA and Horwitz, Rael and Smith, Jennifer E and Jaffe, Jules S},
  journal={Nature Communications},
  volume={7},
  number={1},
  pages={12093},
  year={2016},
  publisher={Nature Publishing Group UK London}
}

@article{shahani2021design,
  title={Design and testing of an underwater microscope with variable objective lens for the study of benthic communities},
  author={Shahani, Kamran and Song, Hong and Mehdi, Syed Raza and Sharma, Awakash and Tunio, Ghulam and Qureshi, Junaidullah and Kalhoro, Noor and Khaskheli, Nooruddin},
  journal={Journal of Marine Science and Application},
  volume={20},
  pages={170--178},
  year={2021},
  publisher={Springer}
}

@inproceedings{mou2022reconfigurablerobots,
  title={Reconfigurable robots for scaling reef restoration},
  author={Mou, Serena and Tsai, Dorian and Dunbabin, Matthew},
  booktitle={IEEE International Conference on Robotics and Automation Workshop on Robots for Climate Change},
  year={2022}
}

@article{pollock2017coral,
  title={Coral larvae for restoration and research: A large-scale method for rearing \emph{Acropora millepora} larvae, inducing settlement, and establishing symbiosis},
  author={Pollock, F Joseph and Katz, Sefano M and van de Water, Jeroen AJM and Davies, Sarah W and Hein, Margaux and Torda, Gergely and Matz, Mikhail V and Beltran, Victor H and Buerger, Patrick and Puill-Stephan, Eneour and others},
  journal={PeerJ},
  volume={5},
  pages={e3732},
  year={2017},
  publisher={PeerJ Inc.}
}

@article{dela2020enhancing,
  title={Enhancing coral recruitment through assisted mass settlement of cultured coral larvae},
  author={Dela Cruz, Dexter W and Harrison, Peter L},
  journal={{PLoS One}},
  volume={15},
  number={11},
  year={2020}
}

@article{raine2025ai,
  title={{AI}-driven Dispensing of Coral Reseeding Devices for Broad-scale Restoration of the {Great Barrier Reef}},
  author={Raine, Scarlett and Moshirian, Benjamin and Fischer, Tobias},
  journal={arXiv preprint arXiv:2509.01019},
  year={2025}
}

@article{severati2024autospawner,
  title={The {AutoSpawner} system: Automated \emph{ex situ} spawning and fertilisation of corals for reef restoration},
  author={Severati, Andrea and Nordborg, F Mikaela and Heyward, Andrew and Wahab, Muhammad A Abdul and Brunner, Christopher A and Montalvo-Proano, Jose and Negri, Andrew P},
  journal={Journal of Environmental Management},
  volume={366},
  pages={121886},
  year={2024},
  publisher={Elsevier}
}

@article{hughes2017global,
  title={Global warming and recurrent mass bleaching of corals},
  author={Hughes, Terry P and Kerry, James T and {\'A}lvarez-Noriega, Mariana and {\'A}lvarez-Romero, Jorge G and Anderson, Kristen D and Baird, Andrew H and Babcock, Russell C and Beger, Maria and Bellwood, David R and Berkelmans, Ray and others},
  journal={Nature},
  volume={543},
  number={7645},
  pages={373--377},
  year={2017},
  publisher={Nature Publishing Group UK London}
}

@article{banaszak2023applying,
  title={Applying coral breeding to reef restoration: Best practices, knowledge gaps, and priority actions in a rapidly-evolving field},
  author={Banaszak, Anastazia T and Marhaver, Kristen L and Miller, Margaret W and Hartmann, Aaron C and Albright, Rebecca and Hagedorn, Mary and Harrison, Peter L and Latijnhouwers, Kelly RW and Mendoza Quiroz, Sandra and Pizarro, Valeria and others},
  journal={Restoration Ecology},
  volume={31},
  number={7},
  year={2023},
  publisher={Wiley Online Library}
}

@article{randall2020sexual,
  title={Sexual production of corals for reef restoration in the {Anthropocene}},
  author={Randall, Carly J and Negri, Andrew P and Quigley, Kate M and Foster, Taryn and Ricardo, Gerard F and Webster, Nicole S and Bay, Line K and Harrison, Peter L and Babcock, Russ C and Heyward, Andrew J},
  journal={Marine Ecology Progress Series},
  volume={635},
  pages={203--232},
  year={2020}
}

@article{morelli2021automating,
  title={Automating cell counting in fluorescent microscopy through deep learning with {c-ResUnet}},
  author={Morelli, Roberto and Clissa, Luca and Amici, Roberto and Cerri, Matteo and Hitrec, Timna and Luppi, Marco and Rinaldi, Lorenzo and Squarcio, Fabio and Zoccoli, Antonio},
  journal={Scientific Reports},
  volume={11},
  number={1},
  pages={22920},
  year={2021}
}

@article{falk2019u,
  title={{U-Net: D}eep learning for cell counting, detection, and morphometry},
  author={Falk, Thorsten and Mai, Dominic and Bensch, Robert and {\c{C}}i{\c{c}}ek, {\"O}zg{\"u}n and Abdulkadir, Ahmed and Marrakchi, Yassine and B{\"o}hm, Anton and Deubner, Jan and J{\"a}ckel, Zoe and Seiwald, Katharina and others},
  journal={Nature Methods},
  volume={16},
  number={1},
  pages={67--70},
  year={2019}
}

@article{weldrick2022promising,
  title={A promising approach to quantifying pteropod eggs using image analysis and machine learning},
  author={Weldrick, Christine K},
  journal={Frontiers in Marine Science},
  volume={9},
  pages={869252},
  year={2022},
  publisher={Frontiers Media SA}
}

@article{zhang2021shrimp,
  title={Shrimp egg counting with fully convolutional regression network and generative adversarial network},
  author={Zhang, Junjie and Yang, Guowei and Sun, Lihui and Zhou, Chao and Zhou, Xuefang and Li, Qian and Bi, Meihua and Guo, Jianlin},
  journal={Aquacultural Engineering},
  volume={94},
  pages={102175},
  year={2021}
}

@article{ditria2021annotated,
  title={Annotated video footage for automated identification and counting of fish in unconstrained seagrass habitats},
  author={Ditria, Ellen M and Connolly, Rod M and Jinks, Eric L and Lopez-Marcano, Sebastian},
  journal={Frontiers in Marine Science},
  volume={8},
  pages={629485},
  year={2021}
}

@article{connolly2021improved,
  title={Improved accuracy for automated counting of a fish in baited underwater videos for stock assessment},
  author={Connolly, Rod M and Fairclough, David V and Jinks, Eric L and Ditria, Ellen M and Jackson, Gary and Lopez-Marcano, Sebastian and Olds, Andrew D and Jinks, Kristin I},
  journal={Frontiers in Marine Science},
  volume={8},
  pages={658135},
  year={2021}
}

@article{garcia2023ten,
  title={Ten years of active learning techniques and object detection: A systematic review},
  author={Garcia, Dibet and Carias, Jo{\~a}o and Ad{\~a}o, Telmo and Jesus, Rui and Cunha, Antonio and Magalh{\~a}es, Luis G},
  journal={Applied Sciences},
  volume={13},
  number={19},
  pages={10667},
  year={2023}
}

@book{Souter2021Status,
  editor    = {Souter, David and Planes, Serge and Wicquart, Jérémy and Logan, Murray and Obura, David and Staub, Francis},
  title     = {Status of Coral Reefs of the World: 2020 Report},
  publisher = {Global Coral Reef Monitoring Network (GCRMN) and International Coral Reef Initiative (ICRI)},
  year      = {2021},
  note      = {{DOI:} 10.59387/WOTJ9184},
  howpublished = {\url{https://gcrmn.net/2020-report/}}
}

@inproceedings{bakana2023digital,
  title={Digital Eye on Endangered Wildlife: Crafting Recognition Datasets through Semi-Automated Annotation},
  author={Bakana, Sibusiso Reuben and Zhang, Yongfei and Yang, Shan},
  booktitle={International Conference on Advances in Image Processing},
  pages={1--7},
  year={2023}
}

@inproceedings{desmond2021semi,
  title={Semi-automated data labeling},
  author={Desmond, Michael and Duesterwald, Evelyn and Brimijoin, Kristina and Brachman, Michelle and Pan, Qian},
  booktitle={NeurIPS 2020 Competition and Demonstration Track},
  pages={156--169},
  year={2021},
}

@inproceedings{dayoub2017episode,
  title={Episode-based active learning with {B}ayesian neural networks},
  author={Dayoub, Feras and Sunderhauf, Niko and Corke, Peter I},
  booktitle={IEEE Conference on Computer Vision and Pattern Recognition Workshops},
  pages={26--28},
  year={2017}
}

@article{norouzzadeh2021deep,
  title={A deep active learning system for species identification and counting in camera trap images},
  author={Norouzzadeh, Mohammad Sadegh and Morris, Dan and Beery, Sara and Joshi, Neel and Jojic, Nebojsa and Clune, Jeff},
  journal={Methods in Ecology and Evolution},
  volume={12},
  number={1},
  pages={150--161},
  year={2021},
  publisher={Wiley Online Library}
}

@article{ben2025benthic,
  title={{The Benthic Underwater Microscope imaging PAM (BUMP): A} non-invasive tool for \emph{in situ} assessment of microstructure and photosynthetic efficiency},
  author={Ben-Zvi, Or and Roberts, Paul and Ratelle, Devin and Snider, Joseph and Lertvilai, Pichaya and Wangpraseurt, Daniel and Deheyn, Dimitri D and Smith, Jennifer E and Jaffe, Jules S},
  journal={Methods in Ecology and Evolution},
  year={2025},
  publisher={Wiley Online Library}
}

\end{document}